\documentclass[pdflatex,sn-mathphys-num, iicol]{sn-jnl}

\usepackage{graphicx}%
\usepackage{multirow}%
\usepackage{amsmath,amssymb,amsfonts}%
\usepackage{amsthm}%
\usepackage{mathrsfs}%
\usepackage[title]{appendix}%
\usepackage{xcolor}%
\usepackage{textcomp}%
\usepackage{manyfoot}%
\usepackage{booktabs}%
\usepackage{algorithm}%
\usepackage{algorithmicx}%
\usepackage{algpseudocode}%
\usepackage{listings}%

\theoremstyle{thmstyleone}%
\theoremstyle{thmstyletwo}%

\theoremstyle{thmstylethree}%

\begin{document}

\title[Article Title]{Toward Generalizable Deep Learning Based Peatland Fire Detection via Walsh–Hadamard Transform and Domain Adaptation}


\author*[1]{\fnm{Emadeldeen} \sur{Hamdan}}\email{ehamda@uic.edu}

\author[2]{\fnm{Ahmad Faiz} \sur{Tharima}}\email{ahmad.faiz@bomba.gov.my}

\author[3]{\fnm{Mohd Zahirasri Mohd} \sur{Tohir}}\email{zahirasri@upm.edu.my}

\author[4]{\fnm{Dayang Nur Sakinah} \sur{Musa}}\email{dns.m@ums.edu.my}

\author[1]{\fnm{Erdem} \sur{Koyuncu}}\email{ekoyuncu@uic.edu}

\author[5]{\fnm{Adam C.} \sur{Watts}}\email{adam.watts@usda.gov}

\author[1]{\fnm{Ahemt Enis} \sur{Cetin}}\email{aecyy@uic.edu}

\affil*[1]{\orgdiv{Department of Electrical and Computer Engineering}, \orgname{University of Illinois Chicago}, \orgaddress{\city{Chicago}, \state{Illinois}, \country{USA}}}

\affil[2]{\orgdiv{Fire and Rescue Department of Malaysia}, \orgaddress{\country{Malaysia}}}

\affil[3]{\orgdiv{Safety Engineering and Reliability (SafER) Research Group, Dept. of Chemical and Environmental Engineering, Faculty of Engineering}, \orgname{Universiti Putra Malaysia}, \orgaddress{\country{Malaysia}}}

\affil[4]{\orgdiv{International Tropical Forestry Programme}, \orgname{Universiti Malaysia Sabah}, \orgaddress{ \country{Malaysia}}}

\affil[5]{\orgdiv{USDA Forest Service Pacific Wildland Fire Sciences Laboratory}, \orgaddress{\state{Washington}, \country{USA}}}

\abstract{Machine learning-based wildfire detection has advanced significantly using deep learning models trained on large wildfire image and video datasets. However, peatland fires exhibit distinct characteristics, including smoldering combustion, low flame intensity, persistent smoke, and subsurface burning, limiting the effectiveness of conventional wildfire detectors. To address these challenges, we propose an efficient deep learning framework for peatland fire detection based on a Walsh--Hadamard Transform enhanced ResNet-50 (WHT-ResNet-50), which improves feature representation while reducing model complexity. To enable efficient deployment, the training-time architecture is structurally reparameterized into an equivalent inference model without sacrificing detection performance. Furthermore, the proposed framework leverages wildfire-to-peatland domain adaptation through transfer learning and introduces a mixed-domain training strategy that produces a unified detector capable of recognizing both wildfire and peatland fire events. Experimental results demonstrate that transfer learning substantially improves peatland fire detection under limited-data conditions, while the proposed WHT-ResNet-50 achieves higher accuracy and F1-score than conventional architectures with fewer parameters. The structurally reparameterized model further reduces inference cost while preserving detection accuracy. Video-based evaluation demonstrates robust performance with low false alarm rates, achieving a 100\% event detection rate across all positive test videos. Overall, the proposed framework provides an accurate, efficient, and practical solution for early peatland fire detection.
}




\maketitle

\section{Introduction}\label{intro}

Wildfires pose a significant threat to ecosystems, infrastructure, and human health worldwide. Recent advances in machine learning, particularly deep learning, have enabled accurate and automated wildfire detection using imagery acquired from surveillance cameras, drones, and satellites \cite{gunay2015real,AslanGTCICASSP19,toreyin2006computer,yuan2010integrated,cetin2016methods,ccelik2007fire}. Modern wildfire detection systems commonly employ convolutional neural networks (CNNs), vision transformers, and video-based architectures trained on large-scale wildfire datasets, achieving high detection accuracy under diverse environmental conditions.

Despite these advances, existing wildfire detectors do not readily generalize to peatland fires \cite{zelioli2025peatland,saleh2025peatland,choy2025tropical}. Unlike conventional wildfires, peatland fires are dominated by low-temperature smoldering combustion, weak or absent visible flames, persistent smoke, and subsurface fire propagation \cite{lara2025peatland,unik2025application}. These characteristics significantly reduce the discriminative visual cues relied upon by conventional deep learning models. Furthermore, the limited availability of labeled peatland fire data makes training robust deep learning models from scratch impractical.

To address these challenges, we propose an efficient deep learning framework for peatland fire detection based on a Walsh--Hadamard Transform enhanced ResNet-50 (WHT-ResNet-50). The proposed architecture incorporates transform-domain feature enhancement to improve feature representation while reducing model complexity. To further improve deployment efficiency, the training-time network is structurally reparameterized into an equivalent inference model, reducing computational overhead without sacrificing detection accuracy~\cite{vasu2023fastvit}. In addition, the framework leverages wildfire-to-peatland domain adaptation through transfer learning and introduces a mixed-domain training strategy that combines wildfire and peatland samples to improve generalization across both fire domains.

The main contributions of this work are summarized as follows:

\begin{itemize}

\item We propose a Walsh--Hadamard Transform enhanced ResNet-50 (WHT-ResNet-50) architecture that improves feature representation while reducing model complexity for peatland fire detection.

\item We employ structural reparameterization to fuse the training-time architecture into an equivalent inference model, enabling faster and more efficient deployment while maintaining detection performance.

\item We investigate wildfire-to-peatland domain adaptation through transfer learning and propose a mixed-domain training strategy that improves peatland fire detection while preserving detection capability on conventional wildfire imagery.

\item Extensive experiments demonstrate that the proposed framework achieves competitive or superior performance compared with conventional CNN and transformer-based architectures using fewer learnable parameters, while providing robust frame-level and video-level detection with low false alarm rates.

\end{itemize}

The proposed framework provides an efficient and practical solution for early peatland fire detection, supporting real-time deployment in data-scarce environments and offering a promising approach for cross-domain wildfire monitoring.

\section{Background}
\label{sec:Backgroud}

\subsection{Deep Learning for Fire and Smoke Detection}

Deep learning, particularly convolutional neural networks (CNNs), has become the dominant approach for automatic fire and smoke detection due to its ability to learn discriminative visual representations directly from data~\cite{toreyin2006computer,choudhry2025comparison}. Architectures such as ResNet~\cite{he2016deep,reddy2025effective} have demonstrated strong performance in wildfire monitoring, offering an effective balance between detection accuracy, computational efficiency, and deployment simplicity.

Beyond conventional 2D CNNs, spatiotemporal models such as 3D CNNs have been explored to capture temporal dynamics across consecutive frames~\cite{jia2026wildfire3d}. Although these approaches can improve motion modeling, they generally require substantially greater computational resources and larger annotated datasets. Consequently, efficient 2D CNNs remain an attractive choice for real-time fire monitoring~\cite{verstockt2013multi,zeng2020coedge,samavat2025early,aslan2019early,aslan2020deep}, particularly in data-scarce scenarios such as peatland fire detection.

\subsection{Domain Adaptation}

Domain adaptation aims to transfer knowledge learned from a source domain to a related target domain while minimizing the amount of labeled target data required. In wildfire monitoring, this is particularly valuable because collecting and annotating fire imagery across different environments is costly and time-consuming~\cite{hong2024wildfire,imran2025fire}.

Rather than training a peatland detector from scratch, wildfire detection models can be adapted to peatland environments by fine-tuning pretrained networks. This enables the model to retain general wildfire representations while learning the distinctive characteristics of peatland fires, including weak flames, persistent smoke, and low-contrast combustion patterns.

In this work, models pretrained on the GWFP (Global Wildfire Prevention) dataset~\cite{hamdan2026large} are adapted to a Malaysian peatland fire dataset through transfer learning and a mixed-domain training strategy. This enables efficient learning under limited-data conditions while improving robustness across both conventional wildfire and peatland fire domains.

\section{Methodology}
\subsection{Walsh--Hadamard Transform}
\label{subsec:HT}

Peatland fire imagery contains weak and diffuse visual cues that are difficult to capture using conventional spatial convolutions alone. To improve feature representation while maintaining low computational complexity, we employ the Walsh--Hadamard Transform (WHT), an orthogonal transform that decomposes signals into binary-valued basis functions~\cite{agaian2011hadamard,pratt1969hadamard}. Owing to its computational efficiency and lightweight implementation, WHT has recently attracted attention for efficient neural network design~\cite{pan2022block}.

The normalized Hadamard matrix of order $2^m$ is defined recursively as
\begin{equation}
H_m = \frac{1}{\sqrt{2}}
\begin{bmatrix}
H_{m-1} & H_{m-1} \\
H_{m-1} & -H_{m-1}
\end{bmatrix}, 
\quad H_0 = [1],
\end{equation}
where $H_m H_m^{T} = I$, indicating orthogonality. 

Unlike conventional convolutional filters, Hadamard-based operations can be efficiently implemented using only additions and subtractions through the fast Walsh--Hadamard transform (FWHT), avoiding costly multiplications. 
In practice, the normalization factor is often omitted for computational efficiency, as it can be absorbed into subsequent learnable layers~\cite{pan2022deep}.

\begin{figure}
    \centering
    
    \includegraphics[width=\linewidth, height =3cm]{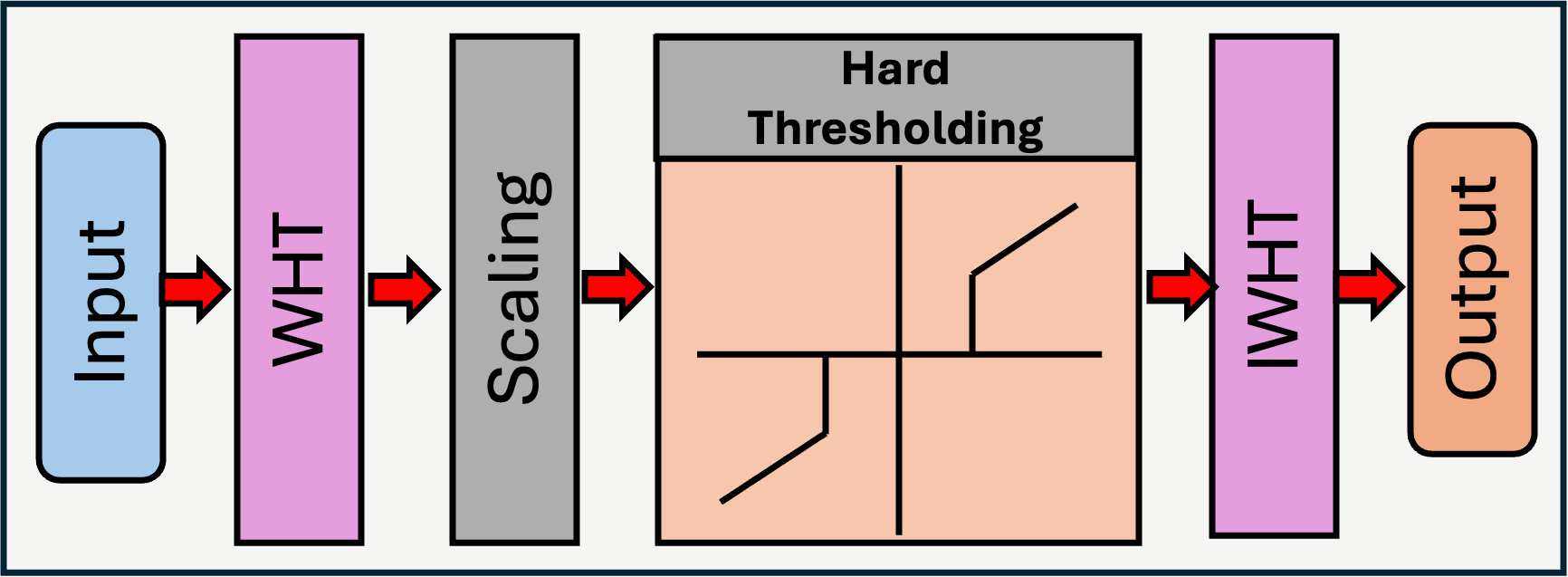}

    \caption{\textbf{Walsh–Hadamard Transform (WHT) layer architecture.} WHT and IWHT denote the forward and inverse Walsh–Hadamard transforms, respectively. The scaling layer performs element-wise multiplication with $N$ trainable parameters followed by a hard-thresholding activation.}
    \label{fig:WHT}
\end{figure}

\begin{figure*}
    \centering
    
    \includegraphics[width=0.9\linewidth, height = 13cm]{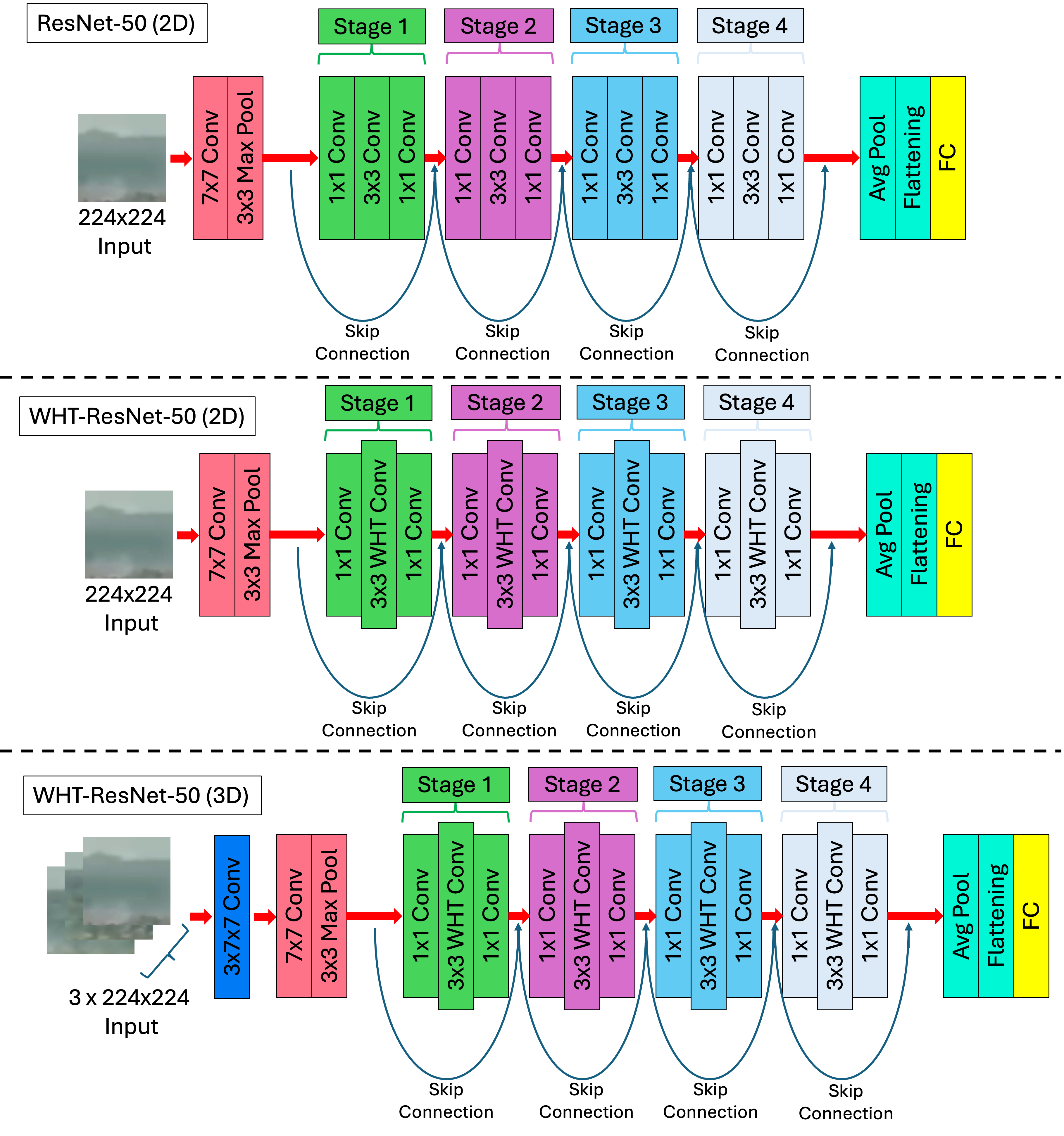}

    \caption{\textbf{Model architectures.} From top to bottom: (1) baseline ResNet-50, (2) the proposed Walsh--Hadamard Transform enhanced ResNet-50 (WHT-ResNet-50), and (3) 3D WHT-ResNet-50. The 2D models operate on a single $224\times224$ image patch extracted from one frame, whereas the 3D model processes three consecutive $224\times224$ patches from adjacent frames. The final fully connected (FC) layer produces two output logits corresponding to the peatland fire and non-fire classes.}
    \label{fig:Models}
\end{figure*}

\subsection{Learnable Thresholding Nonlinearity}

The WHT produces a de-correlated representation in which informative signal components are typically concentrated in a subset of transform coefficients, while noise is distributed among lower-magnitude coefficients. Motivated by this property, we employ a learnable hard-thresholding operation to suppress insignificant coefficients and enhance discriminative features~\cite{zhu2026vhu,donoho1995noising}.

As illustrated in Fig.~\ref{fig:WHT}, feature maps are first transformed into the Hadamard domain, followed by a learnable scaling layer and a hard-thresholding function. The retained coefficients are then reconstructed using the inverse Walsh--Hadamard Transform (IWHT). Although hard thresholding is non-differentiable at the threshold boundary, stable end-to-end optimization is achieved in practice using standard automatic differentiation.

The proposed thresholding operation improves robustness to noise and reduces overfitting on small datasets. As demonstrated in Table~\ref{tab:model_eval}, this leads to more stable training and improved detection performance compared with conventional activation functions used in the baseline ResNet-50.

\subsection{WHT-ResNet-50}

The proposed Walsh--Hadamard Transform enhanced ResNet-50 (WHT-ResNet-50) integrates the proposed WHT layer into the standard residual architecture of ResNet-50~\cite{he2016deep}, as illustrated in Fig.~\ref{fig:Models}. Specifically, the WHT layer replaces the $3\times3$ convolution within the bottleneck residual blocks across Stages 1--4, while the surrounding $1\times1$ convolutions and residual connections are retained. This placement follows the block-level design introduced in Pan et al~\cite{pan2022block}, which demonstrated that incorporating transform-domain processing at these locations provides an effective alternative to conventional spatial convolutions.

Within each WHT layer (Fig.~\ref{fig:WHT}), input feature maps are transformed into the Hadamard domain using the WHT, processed through a learnable scaling layer followed by the proposed hard-thresholding operation, and reconstructed using the inverse Walsh--Hadamard Transform (IWHT). The scaling layer introduces only $N$ trainable parameters, providing a lightweight alternative to conventional convolutional operations~\cite{yuan2023comprehensive}.

The resulting architecture is a hybrid model that combines convolutional feature extraction with efficient transform-domain processing, producing compact feature representations while maintaining the expressive capability of ResNet-50~\cite{pan2022block}. The WHT representation is particularly relevant to fire detection because fire and smoke regions exhibit distinctive spatial intensity variations, irregular boundaries, and heterogeneous texture patterns relative to many background regions. Transforming intermediate feature maps into the Hadamard domain decomposes these spatial responses into orthogonal components representing different patterns of spatial variation. The subsequent learnable scaling operation allows the network to adaptively emphasize transform components that are informative for distinguishing fire-related features, while the proposed hard-thresholding operation suppresses weak or less informative coefficients. Consequently, the WHT layer can selectively retain discriminative structural and intensity variations associated with fire and smoke while reducing redundant feature responses and the number of learnable parameters, making the resulting architecture well suited for efficient peatland fire detection.

\subsection{Structural Reparameterization for Efficient Inference}

To improve deployment efficiency, the proposed WHT-ResNet-50 adopts the structural reparameterization strategy introduced in FastViT~\cite{vasu2023fastvit}. During training, convolutional layers are followed by Batch Normalization (BN). After training, each convolution--BN pair is algebraically fused into an equivalent convolution layer, eliminating BN during inference without altering the learned mapping.

Given a convolution followed by Batch Normalization,

\begin{equation}
y=\gamma
\frac{\mathrm{Conv}(x;W,b)-\mu}
{\sqrt{\sigma^2+\varepsilon}}
+\beta,
\end{equation}
the fused convolution is obtained as

\begin{align}
W' &= \frac{\gamma}{\sqrt{\sigma^2+\varepsilon}}W,\\
b' &= \beta+\frac{\gamma}{\sqrt{\sigma^2+\varepsilon}}(b-\mu),
\end{align}
yielding the equivalent inference model

\begin{equation}
y=\mathrm{Conv}(x;W',b').
\end{equation}

This transformation preserves the network output while removing Batch Normalization operations during inference. For a convolution layer with $C_{\mathrm{out}}$ output channels, the trainable parameters are reduced from

\begin{equation}
C_{\mathrm{out}}C_{\mathrm{in}}K_hK_w+2C_{\mathrm{out}}
\end{equation}
to

\begin{equation}
C_{\mathrm{out}}C_{\mathrm{in}}K_hK_w+C_{\mathrm{out}},
\end{equation}
eliminating one trainable parameter per output channel while reducing memory access and improving inference efficiency.

\section{Experiments}
\subsection{Datasets}

Two datasets are used in this study: the Global Wildfire Prevention (GWFP) dataset~\cite{hamdan2026large} for pretraining and a Malaysian peatland fire dataset for domain adaptation and evaluation.

The GWFP dataset provides approximately 27,500 smoke images and 27,500 negative samples collected under diverse environmental conditions. All images are resized to $224\times224$ prior to training.

The peatland dataset consists of 10 videos containing peatland fire and non-fire scenes. To prevent data leakage, data splitting is performed at the video level rather than the frame or patch level. Five-fold cross-validation is employed, where each fold uses seven videos for training and three videos for validation. Frames are extracted from each video and processed using the block-based strategy described in Section~\ref{sub:PreProcessing}. The resulting dataset contains approximately 2,700 fire and 2,700 negative training samples, together with approximately 500 samples per class for validation in each fold.

\subsection{Data Pre-Processing}
\label{sub:PreProcessing}

Peatland fires typically produce diffuse, low-contrast smoke with ambiguous boundaries, making precise bounding-box or segmentation annotation difficult. Therefore, fire detection is formulated as a patch-based binary classification problem rather than an object detection task.

Each high-resolution frame is partitioned into overlapping spatial regions, as illustrated in Fig.~\ref{fig:fire_detection}. Neighboring $224\times224$ blocks are combined to form overlapping $448\times448$ patches, which are subsequently resized to $224\times224$ before being provided to the network. This overlapping strategy preserves fine-grained smoke patterns while improving spatial coverage and localization without requiring explicit bounding-box annotations.

For an image with dimensions $M_i \times N_i$ and block size $M_b \times N_b$, the number of block rows $R$ and columns $C$ is

\begin{equation}
R=\left\lfloor\frac{M_i}{M_b}\right\rfloor,\qquad
C=\left\lfloor\frac{N_i}{N_b}\right\rfloor.
\end{equation}
where $\lfloor \cdot \rfloor$ denotes the floor function.

To investigate the benefit of temporal information, a 3D input representation is also evaluated. For each spatial location, corresponding patches from three consecutive frames are stacked to form a $3\times224\times224$ input tensor. This enables the network to exploit temporal consistency in smoke evolution while maintaining the same spatial preprocessing pipeline.

\subsection{Training Setup}

The proposed WHT-ResNet-50 is first pretrained on the GWFP dataset and subsequently fine-tuned on the peatland dataset using five-fold cross-validation. Data augmentation includes rotation, flipping, and center-crop.

Training was performed using the Adam optimizer with an initial learning rate of $0.001$, batch size $256$, and for $150$ epochs. The model with the lowest validation loss was selected for evaluation.

All experiments were conducted using an NVIDIA RTX A6000 GPU. Performance is evaluated using Accuracy (ACC), Precision, Recall, and F1-score for patch-level classification. In addition, video-level performance is reported using the True Detection Rate (TDR) and False Alarm Rate (FAR):

\begin{align}
\mathrm{ACC}&=\frac{TP+TN}{TP+TN+FP+FN},\\
\mathrm{Precision}&=\frac{TP}{TP+FP},\\
\mathrm{Recall}&=\frac{TP}{TP+FN},\\
\mathrm{F1}&=\frac{2PR}{P+R}.
\end{align}

\begin{align}
\mathrm{TDR}&=
\frac{\text{Detected Peatland Frames}}
{\text{Peatland Frames}}
\times100,\\
\mathrm{FAR}&=
\frac{\text{False Alarm Frames}}
{\text{Total Frames}}
\times100.
\end{align}

\subsection{Results}
\label{sec:results}

\begin{figure*}
    \centering
    \begin{minipage}{0.49\linewidth}
        \centering
        \includegraphics[width=\linewidth, height = 5.5cm]{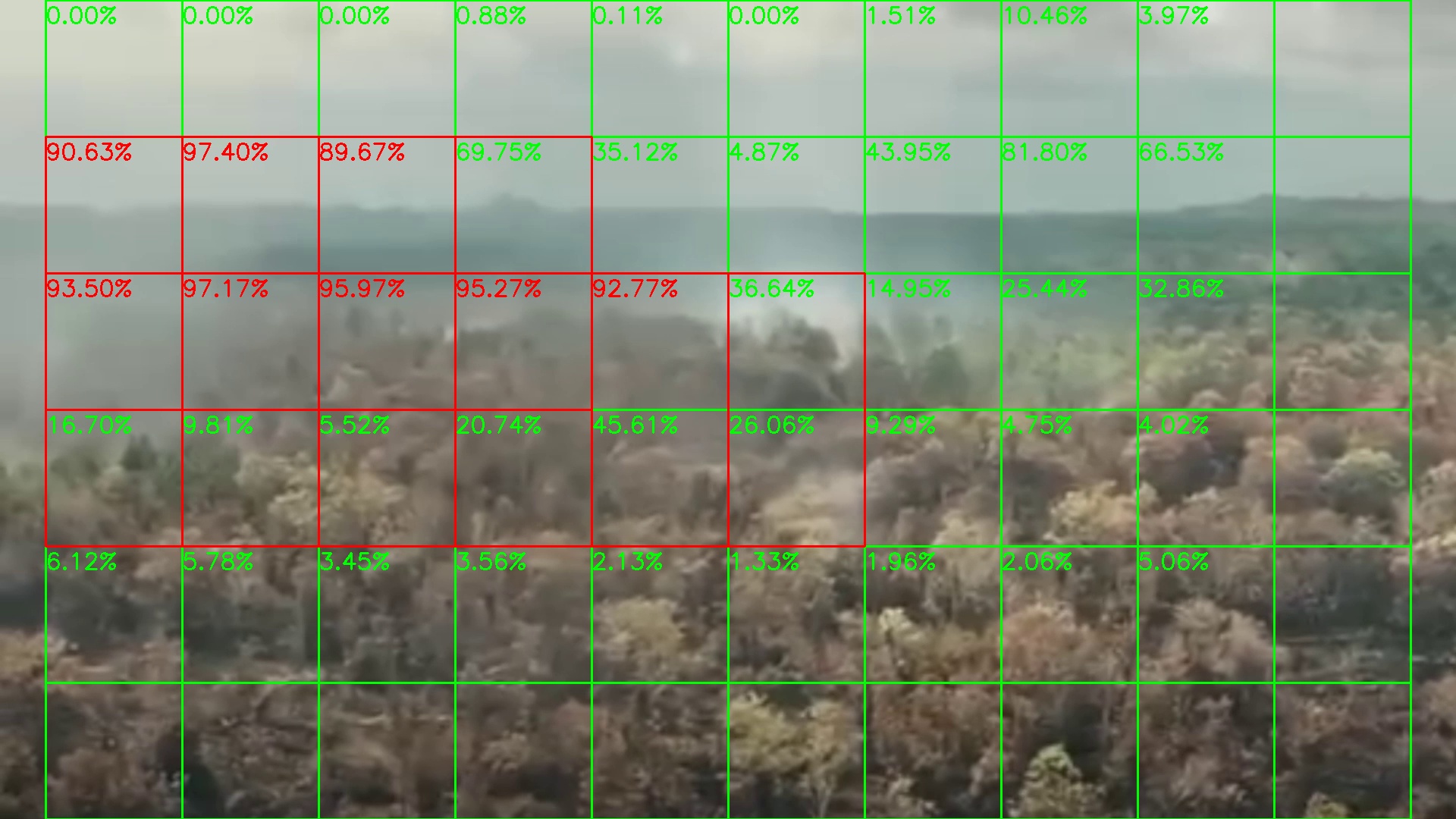}
        \textbf{(A)}
    \end{minipage}
    \hfill
    \begin{minipage}{0.49\linewidth}
        \centering
        \includegraphics[width=\linewidth,height = 5.5cm]{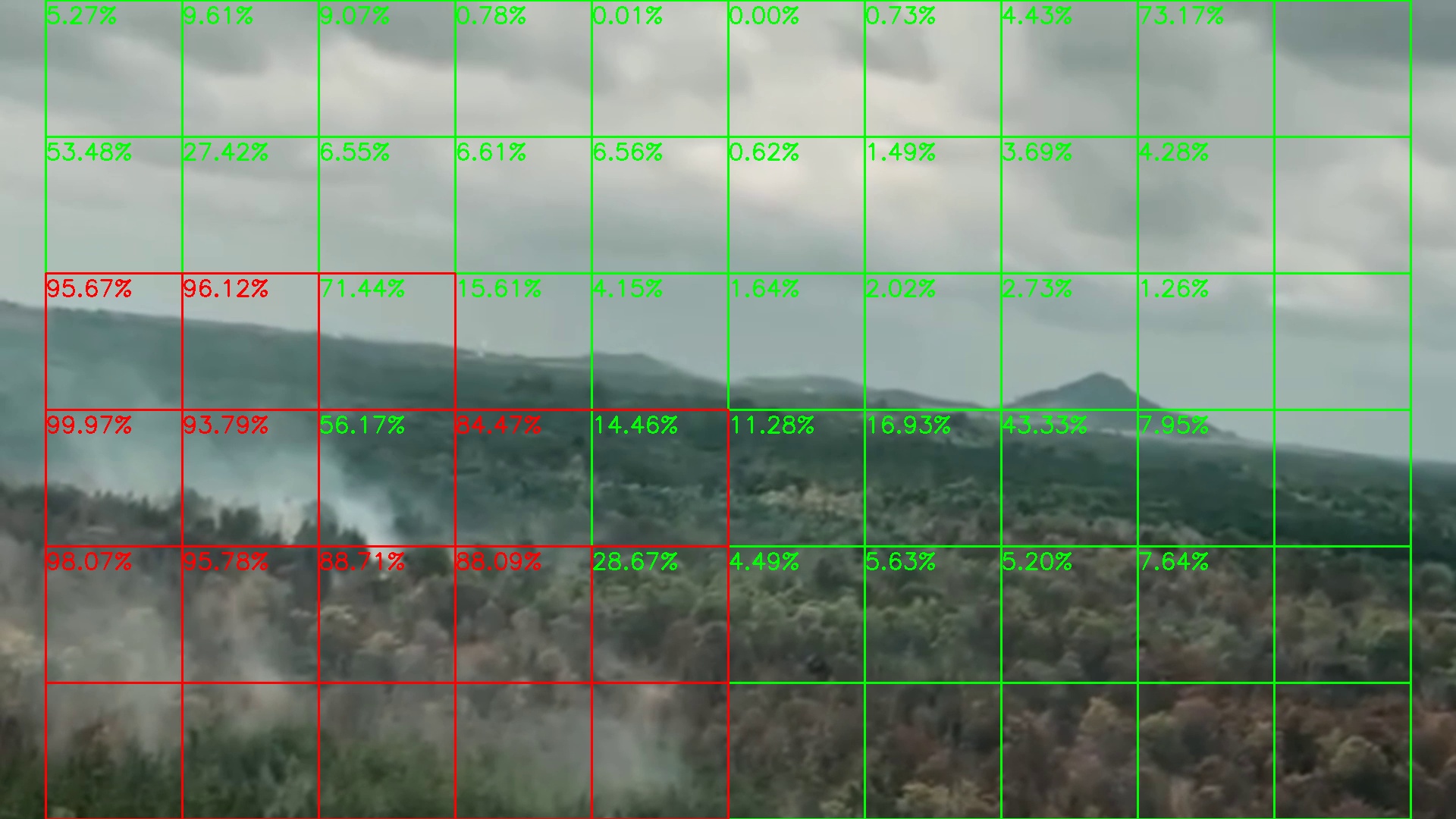}
        \textbf{(B)}
    \end{minipage}

    \caption{\textbf{Peatland Wildfire Smoke Detection Examples:} Two examples of peatland wildfire detection: (A) and (B). Green boxes indicate \emph{no fire/smoke}, while red boxes indicate \emph{fire/smoke}. Numbers inside the boxes denote the predicted fire probability for each patch. Boxes without probabilities or with partial borders occur near frame edges due to the overlapping window scheme.}
    \label{fig:fire_detection}
\end{figure*}

\begin{table*}
\centering
\caption{Model performance (Accuracy, Precision, Recall, and F1 \%) across architectures and the models' number of parameters (No. of Parameters) after 25 training iterations. "Yes" in Pre-Training (PT) column indicates the model was finetuned on the same model trained on GWFP dataset.}
\label{tab:model_eval}
\begin{tabular}{|l|c|c|c|c|c|c|}
\toprule
\textbf{Model} & \textbf{PT}  & \textbf{ACC}& \textbf{Precision}& \textbf{Recall}& \textbf{F1} & \textbf{No. of Param} \\
\midrule
EfficientNet-B5 (2D)~\cite{tan2019efficientnet} & No & 71.2 & 69.8 & 70.4& 70.4&28,36M \\
Swin Transformer (2D)~\cite{SwinTransformer} & No & 72.2 &71.9 & 71.4 & 71.9&27,52M\\
ResNet50 (2D)~\cite{he2016deep}        & No & 71.4 &70.9 & 69.2& 70.1& 23,51M \\
WHT-ResNet-50 (2D)& No & \textbf{77.2} & \textbf{78.2} & \textbf{76.0}& \textbf{77.1}& \textbf{20,58M}\\
\midrule
EfficientNet-B5 (2D)~\cite{tan2019efficientnet} & Yes  & 88.6 & 89.4& 87.0 & 88.2 &28,36M\\
Swin Transformer (2D)~\cite{SwinTransformer} & Yes  & 91.1 &91.9 & 89.7& 90.8&27,52M \\
ResNet50 (2D)~\cite{he2016deep}        & Yes  & 88.9 &90.0 & 86.5& 88.2& 23,51M \\
WHT-ResNet-50 (2D) (ours)& Yes & \textbf{91.6} & \textbf{92.9} & \textbf{90.2}& \textbf{91.5}& \textbf{20,58M}\\
\midrule
WHT-ResNet-50 (3D) (ours) & Yes & 90.0 & 92.0 & 86.5 & 89.2& 20,59M\\
\bottomrule
\end{tabular}
\end{table*}

\begin{table*}[t]
\centering
\caption{Performance evaluation of the proposed peatland wildfire detection model (WHT-ResNet-50) on ten videos, including five negative videos captured from similar peatland environments without peatland fire/smoke occurrences and five positive videos containing peatland fire/smoke events. TDR denotes True Detection Rate, and FAR denotes False Alarm Rate. Event detection measures whether a peatland fire/smoke event was successfully detected within a video. Peatland First Frame represents the ground-truth first occurrence of peatland fire/smoke in the video. No. of Peatland Frames denotes the number of frames containing peatland fire/smoke occurrences.}
\label{tab:peatland_results}

\begin{tabular}{|l| c| c| c| c| c|c|}
\hline
\textbf{Video ID} &
\textbf{Total Frames} &
\textbf{No. of Peatland} &
\textbf{TDR (\%)} &
\textbf{FAR (\%)}
& \textbf{Event Detection}\\
\hline

1 & 732 & 0 & N/A & 0.2 & N/A\\
2 & 120 & 0 & N/A &  4.0 & N/A\\
3 & 180 & 0 & N/A &  8.8 & N/A\\
4 & 180 & 0 & N/A &  2.2 & N/A\\
5 & 270 & 0 & N/A &  3.3 & N/A\\
\hline
6 & 150 & 150 & 100   & 0& 100\% \\
7 & 330 & 330 & 98.4   & 0& 100\% \\
8 & 160 & 160 & 100   & 0 & 100\%\\
9 & 1620 & 1620 & 97.4   & 2.4& 100\% \\
10 & 510 & 510 & 99.50  & 5.8 & 100\%\\

\hline
Average & - & - & 99.1 & 2.6 & 100\%\\
\hline

\end{tabular}
\end{table*}

\begin{table*}[t]
\centering
\caption{Cross-domain transfer learning and retention analysis using the proposed WHT-ResNet-50 model.}
\label{tab:cross_domain_retention}
\renewcommand{\arraystretch}{1.2}
\setlength{\tabcolsep}{6pt}

\begin{tabular}{l c c}
\hline
\textbf{Training Strategy} &
\textbf{Wildfire ACC (\%)} &
\textbf{Peatland ACC (\%)} \\
\hline

Wildfire Only Training & 95.2 & 77.2 \\
Peatland Fine-Tuning Only & 69.0 & 91.6 \\
Mixed-Domain Fine-Tuning & 89.7 & 90.1 \\

\hline
\end{tabular}
\end{table*}

Table~\ref{tab:model_eval} compares the proposed WHT-ResNet-50 with representative convolutional and transformer-based architectures, including ResNet-50~\cite{he2016deep}, EfficientNet-B5~\cite{tan2019efficientnet}, and Swin Transformer~\cite{SwinTransformer}. Each model was evaluated both with and without GWFP pretraining. As expected, pretraining consistently improves performance across all architectures, demonstrating that large-scale wildfire data provides transferable feature representations for peatland fire detection.

Among all evaluated models, the proposed WHT-ResNet-50 achieves the best overall performance, obtaining 91.6\% accuracy and a 91.5\% F1-score after GWFP pretraining. Even without pretraining, WHT-ResNet-50 consistently outperforms the baseline ResNet-50, indicating that the proposed Walsh--Hadamard Transform enhances feature representation and improves discrimination between peatland fire and visually similar background regions.

In addition to its improved detection performance, WHT-ResNet-50 requires fewer learnable parameters (20.6M) than the standard ResNet-50 (23.5M), demonstrating that the proposed architecture achieves higher accuracy with a more compact model. This combination of accuracy and efficiency makes it well suited for deployment in resource-constrained wildfire monitoring systems.

A 3D extension of the proposed architecture, which processes three consecutive frames, was also evaluated. Although temporal information can potentially provide complementary information regarding fire and smoke dynamics, the 3D model achieved slightly lower performance than its 2D counterpart. This does not necessarily indicate that temporal information is ineffective for peatland fire detection. One possible factor is the use of immediately consecutive frames, which can contain highly redundant information at high video frame rates and therefore provide limited additional temporal cues. In addition, non-stationary camera motion can introduce temporal variations unrelated to fire activity, while the current 3D fusion strategy may not optimally distinguish such motion from meaningful fire and smoke dynamics. These observations suggest that the effectiveness of temporal modeling depends not only on the availability of consecutive frames but also on the temporal sampling interval and fusion strategy. Future work will therefore investigate temporally spaced frame sampling, where frames separated by larger intervals are jointly processed, as well as alternative temporal fusion approaches to better exploit meaningful fire and smoke dynamics.

Table~\ref{tab:peatland_results} summarizes the video-level performance of the proposed WHT-ResNet-50. The proposed method achieves an average True Detection Rate (TDR) of 99.1\% while maintaining a low average False Alarm Rate (FAR) of 2.6\%, demonstrating robust performance across diverse peatland monitoring scenarios. In several videos, the proposed system achieved perfect or near-perfect detection with zero false alarms.

Figure~\ref{fig:fire_detection} also illustrates representative failure cases of the proposed method at the patch level, including incorrectly classified background patches and missed fire patches. These errors are primarily observed under challenging cloudy and foggy conditions, where reduced visibility and low contrast make the distinction between fire-related features and the surrounding background more difficult. In particular, fog can produce diffuse visual patterns that resemble smoke, while simultaneously obscuring weak fire signatures, leading to false-positive and false-negative predictions. Furthermore, because detection is performed at the patch level, neighboring patches may contain different proportions of fire and background information, resulting in variations in their predicted confidence values. Nevertheless, the overall low FAR indicates that false detections occur relatively infrequently. These cases highlight adverse weather and visibility conditions as an important challenge for further investigation as additional peatland fire data become available.

Most importantly, the proposed method achieved a 100\% Event Detection Rate by successfully detecting peatland fire/smoke events in all positive videos. From an operational perspective, ensuring that no fire event is completely missed is critical for early-warning and wildfire monitoring systems, making this result particularly significant.

To evaluate knowledge retention during domain adaptation, three training strategies were compared using the proposed WHT-ResNet-50: wildfire-only training, peatland-only fine-tuning, and mixed-domain fine-tuning (Table~\ref{tab:cross_domain_retention}). While fine-tuning substantially improves peatland fire detection, adapting exclusively to the peatland dataset reduces wildfire detection accuracy, indicating catastrophic forgetting of the original domain knowledge.

\begin{table*}[t]
\centering
\caption{Ablation study of structural reparameterization during inference. The full training model is compared with its fused inference counterpart obtained by merging convolution and Batch Normalization layers. Param denotes the number of learnable parameters in millions (M), and Time denotes the average batch inference time in milliseconds (ms).}
\label{tab:ablation}
\setlength{\tabcolsep}{6pt}

\begin{tabular}{lcccccc}
\hline
\textbf{Test Strategy} &
\textbf{ACC} &
\textbf{F1} &
\textbf{Param} 
&
\textbf{Time} 
&
\textbf{FLOPS} 
&
\textbf{Throughput}\\
\hline

WHT-ResNet-50 (Unfused) & 91.0 & 91.1 & 20.58M & 39.42ms & 3.00G & 25.37FPS\\
\textbf{WHT-ResNet-50 (fused)} & \textbf{91.6} & \textbf{91.5} & \textbf{20.55M }& \textbf{38.50ms }& \textbf{2.80G} & \textbf{25.97FPS}\\

\hline
\end{tabular}
\end{table*}

In contrast, mixed-domain fine-tuning preserves high performance on both wildfire and peatland datasets, demonstrating that incorporating representative wildfire samples during adaptation effectively balances knowledge retention and domain-specific learning. These results suggest that mixed-domain adaptation is a practical strategy for improving cross-domain robustness while mitigating catastrophic forgetting.

Moreover, the training and validation learning curves were also examined across the different training strategies. All strategies exhibited stable convergence, while mixed-domain fine-tuning maintained stable validation performance without evidence of optimization instability.

\subsection{Ablation Studies}

Table~\ref{tab:ablation} evaluates the effect of the proposed fused inference strategy based on the structural reparameterization approach adopted from FastViT~\cite{vasu2023fastvit}. Specifically, the WHT-ResNet-50 (Unfused) configuration represents the proposed WHT architecture before structural reparameterization, whereas WHT-ResNet-50 (Fused) represents the same architecture after structural reparameterization and branch fusion for inference. Thus, while the comparison with the standard ResNet-50 in Table~\ref{tab:model_eval} demonstrates the benefit of the proposed WHT-based architecture, the comparison in Table~\ref{tab:ablation} isolates the effect of structural reparameterization.

Replacing the original convolution--Batch Normalization pairs with their fused equivalents preserves the learned network behavior while reducing the number of parameters and inference time. As shown in Table~\ref{tab:ablation}, the fused model achieves slightly higher Accuracy and F1-score, while reducing the model size from 20.58M to 20.55M parameters and decreasing the average inference time from 39.42\,ms to 38.50\,ms per batch. 

Furthermore, the proposed model shows reduced computational cost and achieves approximately 2.5\% higher throughput compared with the corresponding baseline, further supporting its suitability for efficient inference. These results demonstrate that structural reparameterization provides a more efficient deployment model without compromising detection performance.

\section{Conclusion}

This paper presented an efficient WHT-ResNet-50 framework for peatland fire detection that combines transform-domain feature processing, domain adaptation, and structural reparameterization. GWFP pretraining significantly improves peatland fire detection, while mixed-domain adaptation mitigates catastrophic forgetting and preserves performance across both wildfire and peatland domains. The proposed model outperforms representative CNN and transformer-based architectures and achieves a 100\% Event Detection Rate across all positive test videos while maintaining a low False Alarm Rate. Structural reparameterization further improves inference efficiency without compromising detection performance. Future work will investigate deployment on resource-constrained embedded platforms, including FPGA-based systems, with emphasis on real-time performance, memory utilization, and power consumption.

\section*{Acknowledgment}
This work was presented in-part at the 4th International Smoke Symposium (ISS4), Mar. 2026. This work was supported in part by the National Science Foundation (NSF) under Award No. 2531376.
 








\bibliography{sn-bibliography}

@inproceedings{pan2022deep,
  title={Deep neural network with Walsh-Hadamard transform layer for ember detection during a wildfire},
  author={Pan, Hongyi and Badawi, Diaa and Chen, Chang and Watts, Adam and Koyuncu, Erdem and Cetin, Ahmet Enis},
  booktitle={Proceedings of the IEEE/CVF Conference on Computer Vision and Pattern Recognition},
  pages={257--266},
  year={2022}
}

@inproceedings{gunay2015real,
  title={Real-time dynamic texture recognition using random sampling and dimension reduction},
  author={G{\"u}nay, Osman and {\c{C}}etin, A Enis},
  booktitle={Proceedings of the IEEE International Conference on Image Processing},
  series = {ICIP~'15},
  pages={3087--3091},
  year={2015},
  organization={IEEE}
}

@InProceedings{AslanGTCICASSP19,
  author    = {S{\"u}leyman Aslan and
               U\u{g}ur G{\"u}d{\"u}kbay and
               B. U\u{g}ur T{\"o}reyin and
               A. Enis {\c C}etin},
  title     = {Early Wildfire Smoke Detection Based on Motion-based Geometric Image Transformation and Deep Convolutional Generative Adversarial Networks},
  booktitle = {Proceedings of IEEE International Conference on Acoustics, Speech, and Signal Processing},
  series    = {ICASSP '19}, 
  address =   {Brighton, UK},
  publisher = {IEEE},
  month =     {May},
  year      = {2019},
  pages		= {8315-8319},
  bib2html_dl_pdf = "http://www.cs.bilkent.edu.tr/~gudukbay/publications/papers/conf_papers/icassp19.pdf",
  bib2html_pubtype = {Refereed Conference Papers},
  bib2html_rescat = {Computer Vision}
}

@article{hong2024wildfire,
  title={Wildfire detection via transfer learning: a survey},
  author={Hong, Ziliang and Hamdan, Emadeldeen and Zhao, Yifei and Ye, Tianxiao and Pan, Hongyi and Cetin, Ahmet Enis},
  journal={Signal, Image and Video Processing},
  volume={18},
  number={1},
  pages={207--214},
  year={2024},
  publisher={Springer}
}

@inproceedings{he2016deep,
  title={Deep residual learning for image recognition},
  author={He, Kaiming and Zhang, Xiangyu and Ren, Shaoqing and Sun, Jian},
  booktitle={Proceedings of the IEEE Conference on Computer Vision and Pattern Recognition},
  series= {CVPR~'16},
  pages={770--778},
  year={2016}
}

@article{pan2022block,
  title={{Block Walsh--Hadamard} transform-based binary layers in deep neural networks},
  author={Pan, Hongyi and Badawi, Diaa and Cetin, Ahmet Enis},
  journal={ACM Transactions on Embedded Computing Systems},
  volume={21},
  number={6},
  pages={1--25},
  year={2022},
  publisher={ACM New York, NY}
}

@book{cetin2016methods,
  title={Methods and techniques for fire detection: signal, image and video processing perspectives},
  author={Cetin, A Enis and Merci, Bart and G{\"u}nay, Osman and T{\"o}reyin, Beh{\c{c}}et Ugur and Verstockt, Steven},
  year={2016},
  address   = {Cambridge},
  publisher={Academic Press}
}

@inproceedings{ccelik2007fire,
  title={Fire and smoke detection without sensors: Image processing based approach},
  author={{\c{C}}elik, Turgay and {\"O}zkaramanl{\i}, H{\"u}seyin and Demirel, Hasan},
  booktitle={2007 15th European signal processing conference},
  pages={1794--1798},
  year={2007},
  organization={IEEE}
}

@article{yuan2023comprehensive,
  title={A comprehensive review of binary neural network},
  author={Yuan, Chunyu and Agaian, Sos S},
  journal={Artificial Intelligence Review},
  volume={56},
  number={11},
  pages={12949--13013},
  year={2023},
  publisher={Springer}
}

@book{agaian2011hadamard,
  title={Hadamard transforms},
  author={Agaian, Sos S and Sarukhanyan, Hakob and Egiazarian, Karen and Astola, Jaakko},
  year={2011},
  address   = {Bellingham, WA},
  publisher={SPIE Press}
}

@inproceedings{tan2019efficientnet,
  title={Efficientnet: Rethinking model scaling for convolutional neural networks},
  author={Tan, Mingxing and Le, Quoc},
  booktitle={International conference on machine learning},
  pages={6105--6114},
  year={2019},
  organization={PMLR}
}

@INPROCEEDINGS {SwinTransformer,
author = { Liu, Ze and Lin, Yutong and Cao, Yue and Hu, Han and Wei, Yixuan and Zhang, Zheng and Lin, Stephen and Guo, Baining },
booktitle = {Proceedings of the IEEE/CVF International Conference on Computer Vision},
series= {ICCV~'21},
title = {Swin Transformer: Hierarchical Vision Transformer using Shifted Windows},
year = {2021},
pages = {9992-10002},
url = {},
publisher = {IEEE Computer Society},
address = {Los Alamitos, CA, USA},
month =Oct}

@article{yuan2010integrated,
  title={An integrated fire detection and suppression system based on widely available video surveillance},
  author={Yuan, Feiniu},
  journal={Machine Vision and Applications},
  volume={21},
  number={6},
  pages={941--948},
  year={2010},
  publisher={Springer}
}

@article{toreyin2006computer,
  title={Computer vision based method for real-time fire and flame detection},
  author={T{\"o}reyin, B U{\u{g}}ur and Dedeo{\u{g}}lu, Yi{\u{g}}ithan and G{\"u}d{\"u}kbay, U{\u{g}}ur and Cetin, A Enis},
  journal={Pattern recognition letters},
  volume={27},
  number={1},
  pages={49--58},
  year={2006},
  publisher={Elsevier}
}

@inproceedings{choudhry2025comparison,
  title={Comparison of cnn and vision transformers for wildfire detection: A proxy for stubble burning},
  author={Choudhry, Abhimanyu and Jain, Chirag and Singh, Sofia and Ratna, Sanatan},
  booktitle={2025 3rd International Conference on Disruptive Technologies (ICDT)},
  pages={286--289},
  year={2025},
  organization={IEEE}
}

@article{zelioli2025peatland,
  title={Peatland pixel-level classification via multispectral, multiresolution and multisensor data using convolutional neural network},
  author={Zelioli, Luca and Farahnakian, Fahimeh and Middleton, Maarit and Pitk{\"a}nen, Timo P and Tuominen, Sakari and Nevalainen, Paavo and Pohjankukka, Jonne and Heikkonen, Jukka},
  journal={Ecological Informatics},
  volume={90},
  pages={103233},
  year={2025},
  publisher={Elsevier}
}

@article{lara2025peatland,
  title={Peatland fires in Alaska will double by the end of the century},
  author={Lara, Mark Jason and Michaelides, Roger and Anderson, Duncan and Chen, Wenqu and Hall, Emma Catherine and Ludden, Caroline and Schore, Aiden Isaac Gittler and Mishra, Umakant and Scott, Sarah Nicole},
  journal={Scientific reports},
  volume={15},
  number={1},
  pages={35874},
  year={2025},
  publisher={Nature Publishing Group UK London}
}

@article{unik2025application,
  title={Application of Random Forest Algorithm to Analyze the Confidence Level of Forest Fire Hotspots in Riau Peatland.},
  author={Unik, Mitra and Sitanggang, Imas Sukaesih and Syaufina, Lailan and Surati Jaya, I Nengah},
  journal={Journal of Natural Resources \& Environment Management/Jurnal Pengelolaan Sumberdaya Alam dan Lingkungan},
  volume={15},
  number={2},
  year={2025}
}

@article{saleh2025peatland,
  title={Peatland forest monitoring and management solution in Peninsular Malaysia: Optimal parameters for LoRa data},
  author={Saleh, Nur Luqman and Sali, Aduwati and Terng, Liew Jiun and Rahman, Sharifah Mumtazah Syed Ahmad Abdul and Ali, Azizi Mohd and Ali, Borhanuddin Mohd and Razali, Sheriza Mohd and Nuruddin, Ahmad Ainuddin and Ramli, Nordin},
  journal={Ain Shams Engineering Journal},
  volume={16},
  number={6},
  pages={103374},
  year={2025},
  publisher={Elsevier}
}

@article{choy2025tropical,
  title={The tropical peatlands in Indonesia and global environmental change: A multi-dimensional system-based analysis and policy implications},
  author={Choy, Yee Keong and Onuma, Ayumi},
  journal={Regional Science and Environmental Economics},
  volume={2},
  number={3},
  pages={17},
  year={2025},
  publisher={MDPI}
}

@inproceedings{reddy2025effective,
  title={Effective forest fire detection by UAV image using Resnet 50 compared over Densenet},
  author={Reddy, B Tharun Kumar and Sugumar, R},
  booktitle={AIP Conference Proceedings},
  volume={3267},
  number={1},
  pages={020178},
  year={2025},
  organization={AIP Publishing LLC}
}

@article{jia2026wildfire3d,
  title={Wildfire3D: Three-dimensional Wildfire Scene Reconstruction Using Multiview Aerial Imagery},
  author={Jia, Maoqin and Rui, Xue and Zhang, Jun and Song, Weiguo},
  journal={IEEE Transactions on Geoscience and Remote Sensing},
  year={2026},
  publisher={IEEE}
}

@inproceedings{imran2025fire,
  title={Fire Detection and Severity Classification Using Transfer Learning: AI-Driven Recommendations For Wildfire Management},
  author={Imran, Abdullah and Habib, Shaista and Bibi, Shaheen},
  booktitle={2025 6th International Conference on Innovative Computing (ICIC)},
  pages={1--6},
  year={2025},
  organization={IEEE}
}

@article{pratt1969hadamard,
  title={Hadamard transform image coding},
  author={Pratt, William K and Kane, Julius and Andrews, Harry C},
  journal={Proceedings of the IEEE},
  volume={57},
  number={1},
  pages={58--68},
  year={1969},
  publisher={IEEE}
}

@article{zhu2026vhu,
  title={VHU-Net: Variational Hadamard U-Net for Body MRI Bias Field Correction},
  author={Zhu, Xin and Cetin, Ahmet Enis and Durak, Gorkem and Gundogdu, Batuhan and Hong, Ziliang and Pan, Hongyi and Aktas, Ertugrul and Keles, Elif and Savas, Hatice and Oto, Aytekin and others},
  journal={Medical Image Analysis},
  pages={103955},
  year={2026},
  publisher={Elsevier}
}

@article{donoho1995noising,
  title={De-noising by soft-thresholding},
  author={Donoho, David L},
  journal={IEEE transactions on information theory},
  volume={41},
  number={3},
  pages={613--627},
  year={1995},
  publisher={IEEE}
}

@article{verstockt2013multi,
  title={A multi-modal video analysis approach for car park fire detection},
  author={Verstockt, Steven and Van Hoecke, Sofie and Beji, Tarek and Merci, Bart and Gouverneur, Benedict and Cetin, A Enis and De Potter, Pieterjan and Van de Walle, Rik},
  journal={Fire safety journal},
  volume={57},
  pages={44--57},
  year={2013},
  publisher={Elsevier}
}

@article{hamdan2026large,
  title={A Large Scale Open-Source Image and Video Dataset for Robust Wildfire Detection and Classification},
  author={Hamdan, Emadeldeen and Luo, Yingyi and Toreyin, B Ugur and Koyuncu, Erdem and Watts, Adam J and Gudukbay, Ugur and Cetin, Ahmet Enis},
  journal={arXiv preprint arXiv:2606.10174},
  year={2026}
}

@inproceedings{vasu2023fastvit,
  title={Fastvit: A fast hybrid vision transformer using structural reparameterization},
  author={Vasu, Pavan Kumar Anasosalu and Gabriel, James and Zhu, Jeff and Tuzel, Oncel and Ranjan, Anurag},
  booktitle={Proceedings of the IEEE/CVF international conference on computer vision},
  pages={5785--5795},
  year={2023}
}

@article{zeng2020coedge,
  title={CoEdge: Cooperative DNN inference with adaptive workload partitioning over heterogeneous edge devices},
  author={Zeng, Liekang and Chen, Xu and Zhou, Zhi and Yang, Lei and Zhang, Junshan},
  journal={IEEE/ACM Transactions on Networking},
  volume={29},
  number={2},
  pages={595--608},
  year={2020},
  publisher={IEEE}
}

@article{samavat2025early,
  title={Early-stage wildfire detection: a weakly supervised transformer-based approach},
  author={Samavat, Tina and Yazdi, Amirhessam and Yan, Feng and Yang, Lei},
  journal={Fire},
  volume={8},
  number={11},
  pages={413},
  year={2025},
  publisher={MDPI}
}

@inproceedings{aslan2019early,
  title={Early wildfire smoke detection based on motion-based geometric image transformation and deep convolutional generative adversarial networks},
  author={Aslan, S{\"u}leyman and G{\"u}d{\"u}kbay, U{\u{g}}ur and T{\"o}reyin, B U{\u{g}}ur and Cetin, A Enis},
  booktitle={ICASSP 2019-2019 IEEE International Conference on Acoustics, Speech and Signal Processing (ICASSP)},
  pages={8315--8319},
  year={2019},
  organization={IEEE}
}

@inproceedings{aslan2020deep,
  title={Deep convolutional generative adversarial networks for flame detection in video},
  author={Aslan, S{\"u}leyman and G{\"u}d{\"u}kbay, U{\u{g}}ur and T{\"o}reyin, B U{\u{g}}ur and {\c{C}}etin, A Enis},
  booktitle={International Conference on Computational Collective Intelligence},
  pages={807--815},
  year={2020},
  organization={Springer}
}

\end{document}